\documentclass[a4paper]{article}
\usepackage[utf8]{inputenc}
\usepackage{graphicx}
\usepackage{titlesec}
\usepackage{caption}
\usepackage{subcaption}
\usepackage{float}
\usepackage{hyperref}
\usepackage{geometry}
\usepackage{amsmath}
\usepackage{csquotes}
\usepackage{amssymb}
\newgeometry{vmargin={25mm,50mm}, hmargin={35mm,35mm}}
\titleformat{\paragraph}
{\normalfont\normalsize\bfseries}{\theparagraph}{1em}{}
\titlespacing*{\paragraph}
{0pt}{3.25ex plus 1ex minus .2ex}{1.5ex plus .2ex}

\usepackage{fancyhdr}

\fancypagestyle{plain}{
\fancyhf{}
\pagestyle{fancy}
\fancyhead[C]{\includegraphics[height=2cm]{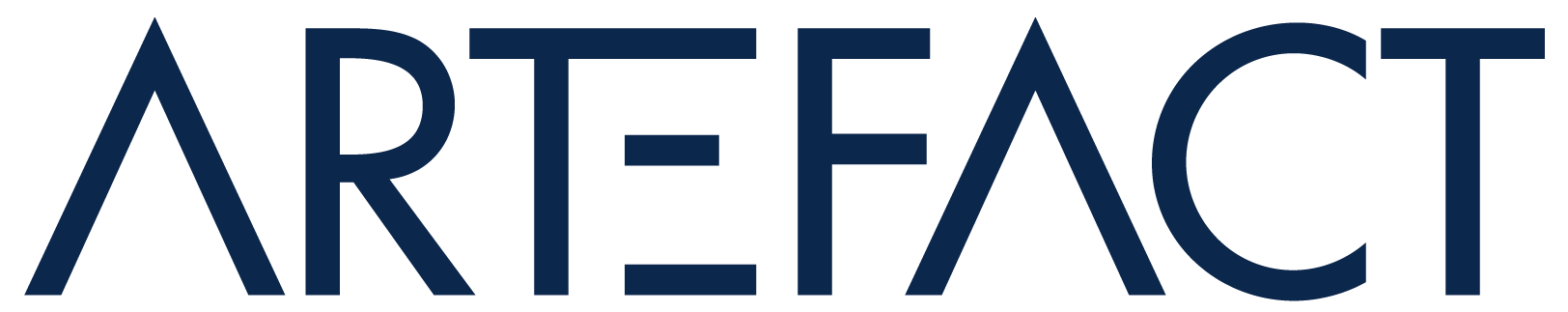}}
\fancyfoot[L]{Copyright \textcopyright \vspace{6pt} 2023 Artefact}
}
\usepackage{amsthm}

\fancypagestyle{artefact}{
\fancyhf{}
\pagestyle{fancy}
\fancyhead[L]{\includegraphics[height=1cm]{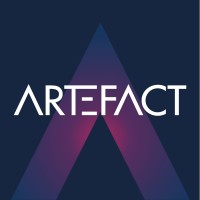}}
\fancyfoot[L]{Copyright \textcopyright \vspace{6pt} 2023 Artefact}
\fancyfoot[R]{ \thepage}
}
\usepackage{authblk}

\author{  
  Marcel Marais \and
  M\'{a}t\'{e} Hartstein\and
  George \v{C}evora \thanks{Correspondence: \texttt{george.cevora@artefact.com}}
  }
\affil{Artefact Ltd, \\ 6th Floor, 51 Eastcheap, London EC3M 1DT}
\date{}

\title{Using linear initialisation to improve speed of convergence and fully-trained error in Autoencoders}

\begin{document}

\maketitle
\begin{abstract}
Good weight initialisation is an important step in successful training of Artificial Neural Networks \cite{narkhede2022review,Hinton}. Over time a number of improvements have been proposed to this process \cite{glorot, He}. In this paper we introduce a novel weight initialisation technique called the \emph{Straddled Matrix Initialiser}. This initialisation technique is motivated by our assumption that major, global-scale relationships in data are linear with only smaller effects requiring complex non-linearities. Combination of Straddled Matrix and ReLU activation function initialises a Neural Network as a de facto linear model, which we postulate should be a better starting point for optimisation given our assumptions. We test this by training autoencoders on three datasets using Straddled Matrix and seven other state-of-the-art weight initialisation techniques. In all our experiments the Straddeled Matrix Initialiser clearly outperforms all other methods.
\end{abstract}

Modern autoencoders are artificial neural networks (ANNs) that are mainly used for dimensionality reduction \cite{Hinton,fournier2019empirical} or anomaly detection \cite{chandola2009anomaly,an2015variational}. Autoencoders consist of a series of hidden layers that transform the input data. The objective of the transformation being that its output matches input as closely as possible, i.e. achieve low reconstruction error \cite{bank2020autoencoders}. 

A major research topic in ANNs is weight initialisation \cite{narkhede2022review}.
In the specific area of autoencoders layer-wise pre-training has been recognised as an improved weight initialisation method \cite{Hinton}, but with the introduction of Glorot, and He initialisation techniques designed for other types of ANNs \cite{glorot,He}, it has became obsolete. The motivation for the development of these very techniques is that the optimal initialisation is dependent on the architecture of the ANN \cite{He}. However, to our knowledge there has not been any novel research into weight initialisation in the context of autonecoders, an omission we address here.

Many problems solved by autoencoders are to a large degree linear. Consequently, we put forward that initialising autoencoders to perform the linear transformation of variables would be a better starting point for gradient descent algorithms than highly nonlinear transformation resulting from commonly used initialisation techniques such as Glorot, and He \cite{glorot,He}. Furthermore, we postulate that the interactions between individual variables do not have larger role than the variables themselves. This lead us to conclude that a good starting point for gradient descent should minimise the overlap of variables, while all variables should have the same weight. When considering ReLU-based autoencoders with the same number of units in all layers these two conclusions lead to initialisation using an Identity matrix, which indeed in this specific case lead to perfect reconstruction without any optimisation of weights. However, this type of autoencoder has only very limited uses, such as Fair Adversarial Networks \cite{cevora2020fair}, as it doesn't establish useful encodings, but merely copies input through a series of identities. General autoencoder architectures utilise non-square weight matrices, which precludes the usage of the Identity matrix. We, therefore, introduce \emph{Straddled Matrix}, an initialisation technique that achieves our aims generalized to any autoencoder architecture.

\begin{figure}[h]
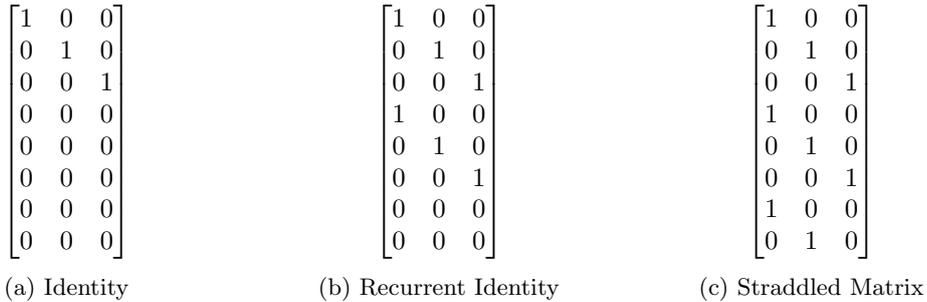

     \centering
    \begin{subfigure}[b]{0.3\textwidth}
             \centering
                               $\begin{bmatrix}
    1 & 0 & 0\\
    0 & 1 & 0\\
    0 & 0 & 1 \\
    0 & 0 & 0 \\
    0 & 0 & 0 \\
    0 & 0 & 0 \\
    0 & 0 & 0 \\
    0 & 0 & 0 \\
    \end{bmatrix}$
             \caption{Identity}
             \label{fig:Identity}
         \end{subfigure}
     \hfill
     \begin{subfigure}[b]{0.3\textwidth}
         \centering
                 $\begin{bmatrix}
1 & 0 & 0\\
0 & 1 & 0\\
0 & 0 & 1 \\
1 & 0 & 0 \\
0 & 1 & 0 \\
0 & 0 & 1 \\
0 & 0 & 0 \\
0 & 0 & 0 \\
\end{bmatrix}$
         \caption{Recurrent Identity }
         \label{fig:recurrent}
     \end{subfigure}
     \hfill
          \begin{subfigure}[b]{0.3\textwidth}
         \centering
         $\begin{bmatrix}
1 & 0 & 0\\
0 & 1 & 0\\
0 & 0 & 1 \\
1 & 0 & 0 \\
0 & 1 & 0 \\
0 & 0 & 1 \\
1 & 0 & 0 \\
0 & 1 & 0 \\
\end{bmatrix}$
         \caption{Straddled Matrix}
         \label{fig:Straddled}
     \end{subfigure}
        \caption{Examples of the three possible ways to initialise a non-square weight matrix with identity.}
        \label{fig:threeInitialisations}
\end{figure}

There are fundamentally three ways that the Identity matrix can be adapted to non-square matrices. First, padding with zeroes as shown in Figure \ref{fig:Identity} and implemented in the current version of TensorFlow (v2.10.0). Second, the Recurrent Identity which repeats the Identity matrix as many times as it will fit, then pads the rest with zeroes as shown in Figure \ref{fig:recurrent}. Recurrent Identity was an undocumented feature of Keras removed by pull request \href{https://github.com/keras-team/keras/pull/11887}{\#11887} and is the most similar initialisation method to the Straddled Matrix which we propose in this paper (Figure \ref{fig:Straddled}). Rationale behind implementation of Recurrent Identity into Keras is unknown to us, while its removal seems to lack any reasoning besides discrepancy with TensorFlow.

\subsection*{Straddled Matrix}
\label{sec:Straddled}
Straddled Matrix initialisation diagonally fills ones for all rows in a repeated manner as shown in Figure \ref{fig:Straddled}, thus, unlike Identity or Recurrent Identity, there is no zero padding. Formally, 
 for a matrix $W_{m \times n}$, Recurrent Identity is equal to the Straddled Matrix if and only if $m \geq n$ and $m$ is a multiple of $n$.
For an $m \times n$ matrix if $m = n$ (a square matrix) the Straddled Matrix is the same as the Identity $I_n$ or $I_m$. However, in the case of a rectangular matrix, specifically when $m > n$, the diagonal of ones continues on the next row, resulting in a single element of the matrix being one for each row  as shown in Figure \ref{fig:53Straddled}. In the $m < n$ case we simply pad the extra columns with zeros as shown in Figure \ref{fig:35Straddled}.  It should be clarified that ${\text{Straddled}}_{3\times5} \neq \text{Straddled}_{5\times3}^T $ neither  $||{\text{Straddled}}_{3\times5} ||\neq ||\text{Straddled}_{5\times3} ||$ as shown in Figure \ref{fig:35v53}.

\begin{figure}[h]
     \centering
     \begin{subfigure}[b]{0.45\textwidth}
         \centering
      $  \begin{bmatrix}
1 & 0 & 0\\
0 & 1 & 0\\
0 & 0 & 1 \\
1 & 0 & 0 \\
0 & 1 & 0 \\
\end{bmatrix}$
         \caption{5x3 Straddled Matrix}
         \label{fig:53Straddled}
     \end{subfigure}
     \hfill
     \begin{subfigure}[b]{0.45\textwidth}
         \centering
       $ \begin{bmatrix}
1 & 0 & 0 & 0 & 0\\
0 & 1 & 0 & 0 & 0\\
0 & 0 & 1 & 0 & 0\\
\end{bmatrix}$
         \caption{3x5 Straddled Matrix}
         \label{fig:35Straddled}
     \end{subfigure}
    
        \caption{The difference between a 3x5 and 5x3 Straddled Matrix}
        \label{fig:35v53}
\end{figure}

The result of the Straddled Matrix initialisation is that every feature receives an equal weight, regardless of the the size of the next hidden layer. When using the ReLU activation function and min-max scaling of features the first forward pass of the model will have no non-linearities and no interactions between features.\\

\section*{Methods}
To compare the performance of various initialisation techniques we test all initialisers using the same autoencoder architecture on three distinct tasks. The structure of the autoencoder was kept as simple as possible to eliminate confounding effects of the architecture. To maximise the difference between Identity, Recurrent Identity and Straddled Matrix we chose to use in the contractive layer to have $50\%+1$ the number of neurons of the preceding layer. The full architecture is described in Figure \ref{figure:architecture}. Root Mean Squared Error (RMSE) of the reconstructed input with respect to the original input was used as a cost function with standard gradient descent as an optimiser.

\begin{figure}[h]
\centering

\begin{tabular}{p{2cm}|p{2.2cm}|p{1.4cm}|p{1.4cm}|p{1.4cm}|p{2.6cm}}
\hline
& Input Layer       & Layer 1  & Layer 2  & Layer 3 & Output Layer      
\\\hline
Size& Dataset \mbox{dependent} & 64 & 33 & 64& Dataset \mbox{dependent} \\
Activation function& N/A & ReLU & ReLU  & ReLU& Sigmoid\\
\end{tabular}
\caption{The architecture of autoencoder used for comparison of different weight initialisation approaches.}
\label{figure:architecture}
\end{figure}

\begin{figure}[h]

\centering
\begin{tabular}{p{0.14\linewidth}|p{0.12\textwidth}|p{0.12\textwidth}|p{0.12\textwidth}|p{0.12\textwidth}|p{0.12\textwidth}}
\hline
{Dataset}         
& {Train/test split} (\%) &{Batch size} & {Num. of epochs} & {Learning rate} & {Num. of runs}\\\hline
Synthetic       & 80/20                   & Full batch & 1000                 & 0.1           & 10          \\
MNIST           & 80/20                   & 256        & 1000                 & 0.1           & 10           \\ 
Swarm Behaviour & 80/20                   & Full batch & 1500                & 0.1           & 10         \\  
\end{tabular}

\caption{The hyper-parameters used for training during the experiemnts}
\label{figure:training}
\end{figure}

To assess the performance  we ran 8 different initialisers listed in Figure \ref{figure:initialisers} on three different datasets described in detail in Section Data \& Results, while keeping the autoencoder architecture constant as per Figure \ref{figure:architecture}. To ensure results are comparable across initialisers all hyper-parameters have been kept the same and in order to mitigate the variance introduced by the stochastic initialisers we ran each experiment multiple times to produce confidence intervals. To give every initialiser a fair chance at convergence they have all been trained for the same number of epochs - leading to overtraining in some instances. The setup for each experiment is shown in Figure \ref{figure:training}.

\begin{figure}[h]
\centering
\begin{tabular}{l|p{10cm}}
\hline
  Initialiser &  Notes \\
\hline
    Straddled &       Novel initialisation proposed in this paper shown in Figure~\ref{fig:Straddled}.  \\
Glorotuniform &          $W\sim U\big(-\sqrt{6/(i+o)},\sqrt{6/(i+o)}\big)$    \\
 Glorotnormal &      $W\sim N\big(0,\sqrt{2/(i+o)}\big)$        \\
     Identity &        Zero padded Identity matrix as shown in Figure \ref{fig:Identity}.      \\
     Henormal &         $W\sim N\big(0,\sqrt{2/i}\big)$     \\
    Heuniform &          $W\sim U\big(-\sqrt{6/i},\sqrt{6/i}\big)$     \\
   Orthogonal &            Orthogonal matrix obtained from the QR decomposition of a matrix of random numbers drawn from a normal distribution.  (TensorFlow 2.11.0 documentation) \\
       Random &          Default  TensorFlow 2.11.0 initialiser sampling from a normal distribution.   \\

\end{tabular}
\caption{The weight initialisers used for performance comparison. $i$ and $o$ refer respectively to the number of inputs and outputs to a layer. }
\label{figure:initialisers}
\end{figure}

To ensure consistency when comparing the performance of the autoencoder using different initialisers we have a defined a condition that if met will be considered the \emph{convergence} point for that initialiser.
At epoch $t$ with loss $l_t$ in the loss curve for the autoencoder with a given intialiser. The autoencoder is said to be converged at $t$ if the loss at every epoch between $t$ and $t + \alpha$ is bounded above by $l_t + \epsilon$ and below by $l_t - \epsilon$. The parameters were selected for each dataset to provide meaningful representation of convergence, the specific values are provided alongside the relevant results.

 The loss values used are epoch-level means taken across the number of runs specified in Figure \ref{figure:training}.

\section*{Data \& Results}
\label{sec:data}
To benchmark the performance of the autoencoder using the Straddled Matrix initialiser against the other initialisers we tested all initialisers on a variety of datasets. Both synthetic and real world datasets with different amounts of features and observations as well as varying levels of non-linearity were used to ensure a holistic overview of the benefits and drawbacks of the different initialisers.

As a preprocessing step all features were scaled to a range $[0,1]$, effectively making the reconstruction error a percentage and thus comparable across datasets.\newpage

\subsubsection*{Synthetic Data}

We used a synthetic dataset with 100 features and 5000 records. Each record was created from a randomly initialised vector of dimension 20, which was transformed using linear and non-linear functions to create a vector of dimension 100 with varying amounts of linearity between elements. The goal was to create a dataset that encodes well and benefits the Straddled Matrix initialisation, which uses the linear transformation of variables as the starting point. The best description of the dataset is the code used to generate it, provided in our public repository.

\begin{figure}[hb!]
\begin{subfigure}{\textwidth}
\centering
    \includegraphics[width=0.9\textwidth]{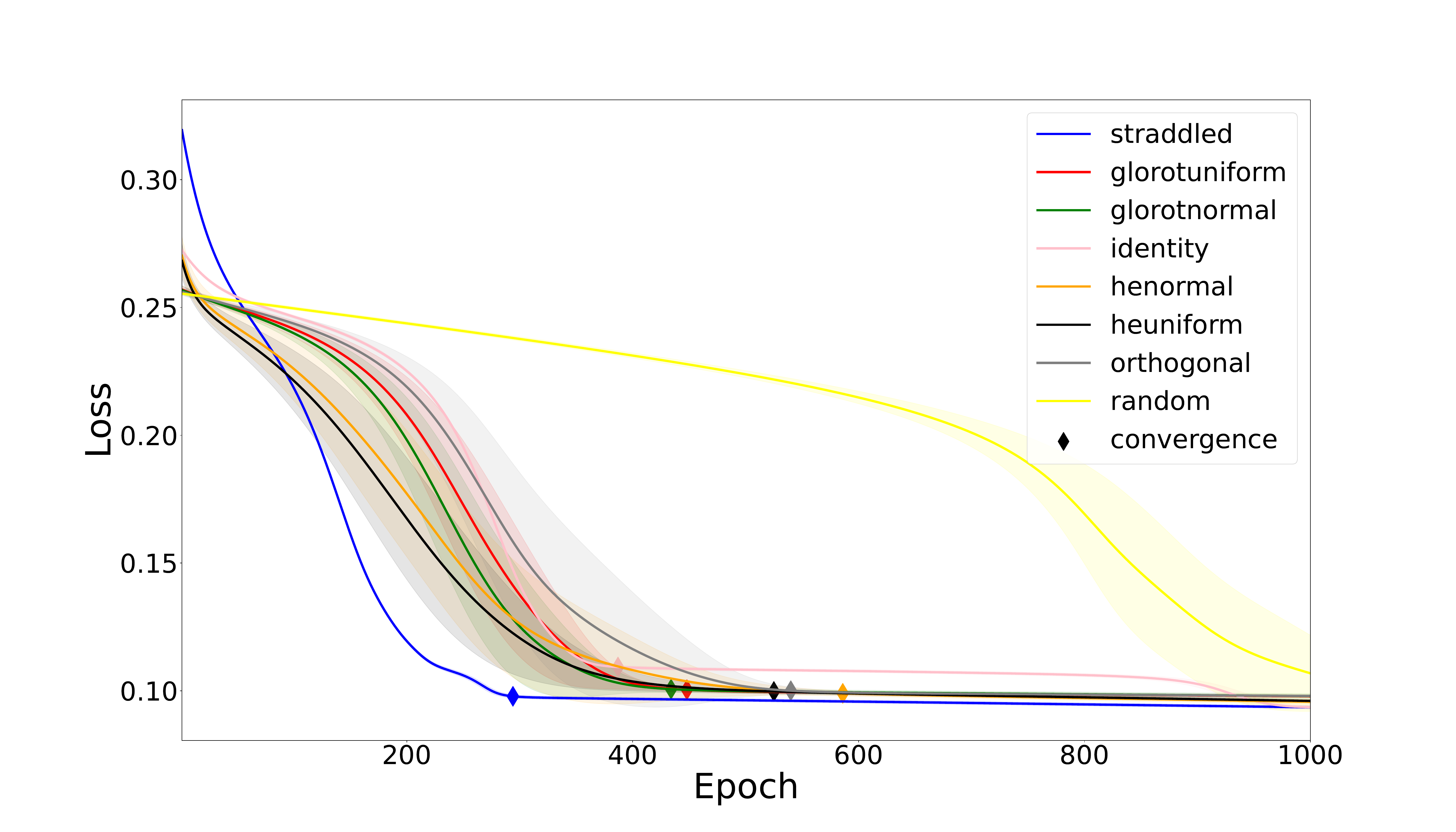}
\caption{Loss as a function of training epoch for each initialiser. The loss values are the average of 10 runs with the associated confidence intervals shown by the semi-transparent areas. Both convergence speed and quality are markedly improved by using Straddled Matrix as opposed to other initialisers.}
\label{fig:allInitSynthetic}
\end{subfigure}
\begin{subfigure}{\textwidth}
\centering
\vspace{10pt}
\begin{tabular}{l|l|l|l}
\hline
  Initialiser &  Converged Epochs &  Converged Loss & p-value\\
\hline
    Straddled &             294.0 &        0.097866 &  \\
Glorotuniform &             448.0 &        0.100448  & $<0.001$\\
 Glorotnormal &             434.0 &        0.100689  & $<0.001$\\
     Identity &             387.0 &        0.109341  & $0.924$\\
     Henormal &             586.0 &        0.099029  & $<0.001$\\
    Heuniform &             525.0 &        0.099770  & $<0.001$\\
   Orthogonal &             540.0 &        0.100072  & $<0.001$\\
       Random  & N/A & N/A & $0.059$\\
\end{tabular}
\caption{Convergence metrics reported as averages across 10 runs, with convergence criteria of $\epsilon = 0.001$, and $\alpha = 100$, with N/A recorded when they were not met throughout the training. The p-values reported correspond to the results of single tailed t-test on the loss at the final epoch of the training across the ten runs. Straddled Matrix converged faster and reached a lower converged loss than all other initialisers.}
\label{figure:allInitSynthetic2}
\end{subfigure}
\caption{Performance of various initialisers on the synthetic dataset.}
\end{figure}

As shown in Figure \ref{fig:allInitSynthetic}, the Straddled Matrix initialiser showed quicker convergence and a slightly lower loss than all other initialisers we tested against on the synthetic dataset. Five out of the seven comparisons between losses reached at the end of the learning came out highly significantly better, and for all seven comparisons the convergence was best and fastest for the Straddled Matrix initialiser as reported in Figure \ref{figure:allInitSynthetic2}.

\newpage

\subsubsection*{MNIST}
The MNIST dataset is a collection of 70,000 handwritten digits in the form of 28x28 greyscale images. The original split contains 60,000 examples in the training set and 10,000 in the the test set. The images were flattened into vectors of size 784, and normalised to be between 0 and 1 - this preprocessing was replicated from a Keras autoencoder blog post \cite{kerasAEBlog}.

\begin{figure}[hb!]
\begin{subfigure}{\textwidth}
\centering
    \includegraphics[width=0.9\textwidth]{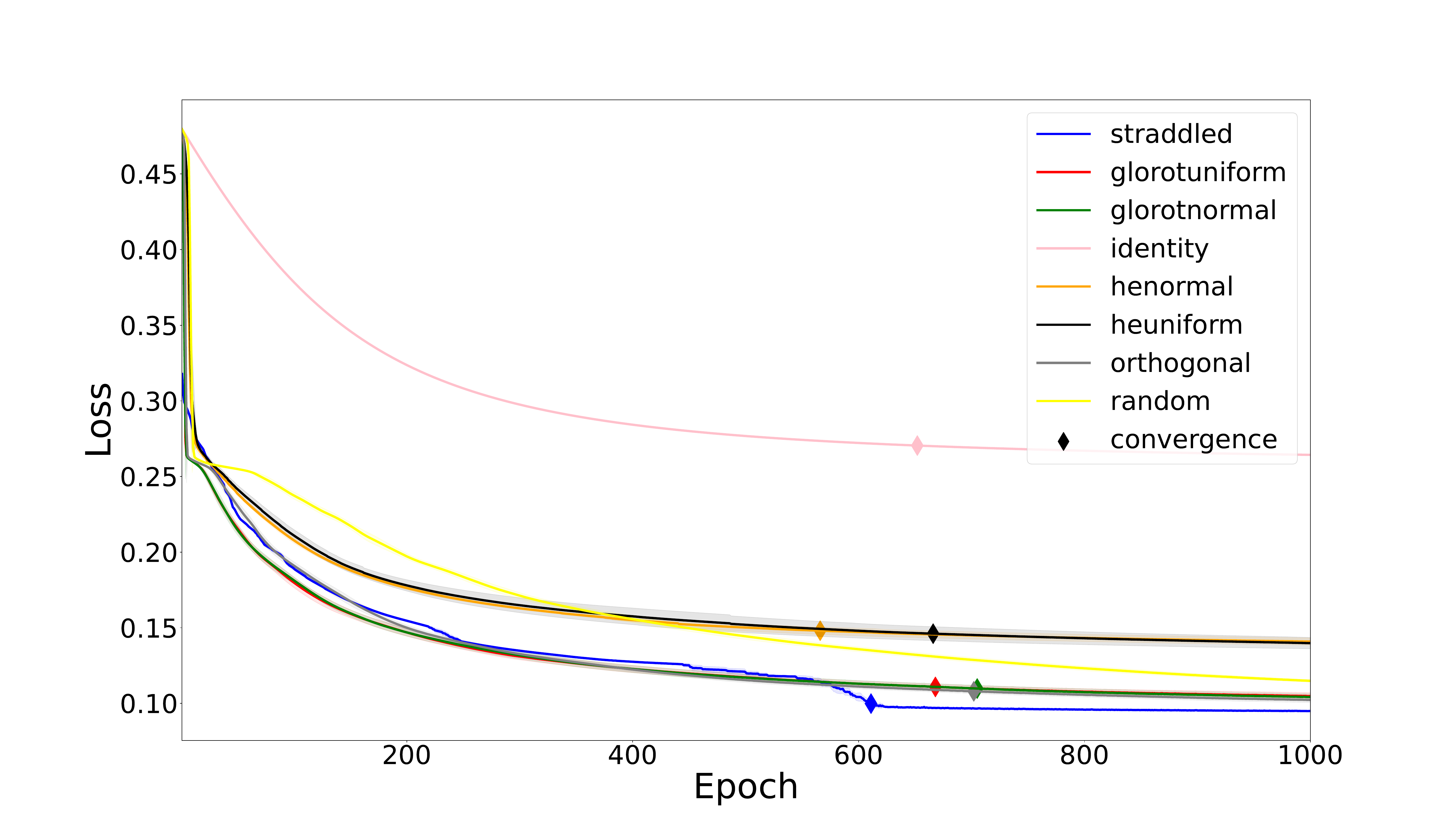}
\caption{Loss as a function of training epoch for each initialiser. The loss values are the average of 10 runs with the associated confidence intervals shown by the semi-transparent areas. Both convergence speed and quality were best for Straddled Matrix initialisation.}
\label{fig:MNISTAll}
\end{subfigure}
\begin{subfigure}{\textwidth}
\centering
\vspace{10pt}
\begin{tabular}{l|l|l|l}
\hline
  Initialiser &  Converged Epochs &  Converged Loss & p-value\\
\hline
    Straddled &             611.0 &        0.099854 &  \\
Glorotuniform &             668.0 &        0.110882  & $<0.001$\\
 Glorotnormal &             705.0 &        0.109809  & $<0.001$\\
     Identity &             652.0 &        0.270451  & $<0.001$\\
     Henormal &             566.0 &        0.148061  &$<0.001$\\
    Heuniform &             666.0 &        0.146084  & $<0.001$\\
   Orthogonal &             702.0 &        0.107960  & $<0.001$\\
       Random &               N/A &             N/A & $<0.001$\\
\end{tabular}
\caption{Convergence metrics reported as averages across 10 runs, with  with convergence criteria of $\epsilon = 0.005$, and $\alpha = 250$, with N/A recorded when they were not met throughout the training. The p-values reported correspond to the results of single tailed t-test on the loss at the final epoch of the training across the ten runs. Straddled Matrix reached a lower converged loss than all other initialisers.}
\label{figure:allInitialiserMNIST}
\end{subfigure}
\caption{Performance of various initialisers on the MNIST dataset.}
\end{figure}

It was expected that the high amount of of non-linearity in the MNIST dataset would have a negative impact on the performance of the Straddled Matrix. However, as shown in Figure \ref{fig:MNISTAll} Straddled Matrix initialisation performed comparably to the best initialisers for the first 500 epochs, and actually converged earlier and at a lower loss than the other three best-performing methods as reported in Figure \ref{figure:allInitialiserMNIST}. The improvement in the final loss for Straddled Matrix initialisation was also confirmed to high significance.

\newpage

\subsubsection*{Swarm Behaviour}
The Swarm Behaviour dataset was constructed by running a survey of human perception of simulated swarming behaviour. The participants were asked to rate to what extent the behaviour could be considered flocking, aligned and grouped. The features of this dataset relate to the position, velocity and other movement properties of each entity in the swarm \cite{swarmBehaviour}.

\begin{figure}[hb!]
\begin{subfigure}{\textwidth}
\centering
    \includegraphics[width=0.9\textwidth]{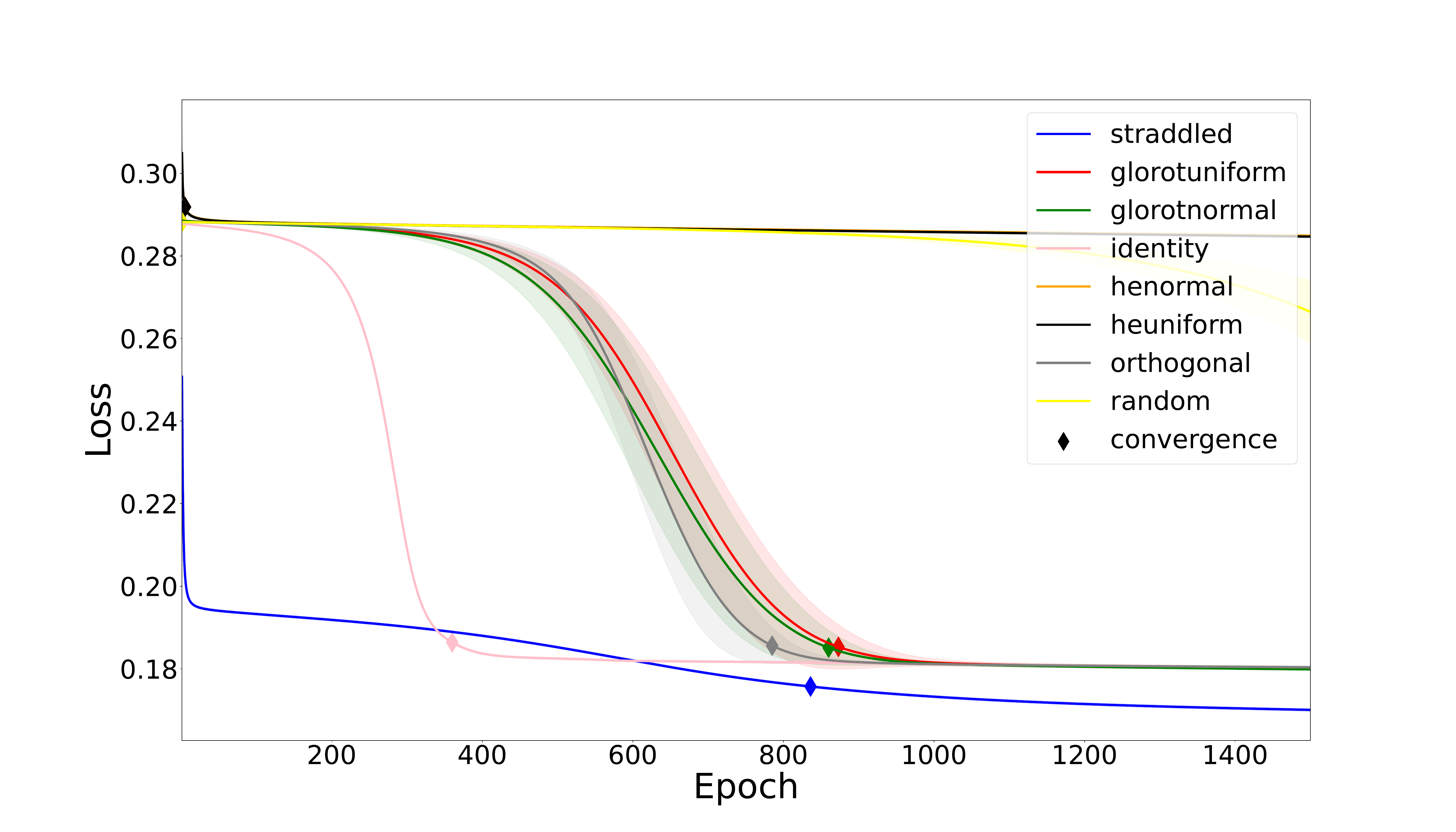}
\caption{Loss as a function of training epoch for each initialiser. The loss values are the average of 10 runs with the associated confidence intervals shown by the semi-transparent areas. Both convergence speed and quality are markedly improved by using Straddled Matrix as opposed to other initialisers. While Identity initialisation converged faster, the loss of converged state was worse.}
\label{fig:SwarmAll}
\end{subfigure}
\begin{subfigure}{\textwidth}
\centering
\vspace{10pt}
\begin{tabular}{l|l|l|l}
\hline
  Initialiser &  Converged Epochs &  Converged Loss & p-value\\
\hline
    Straddled &               836 &        0.175701 &\\
Glorotuniform &               873 &        0.185304 & $<0.001$ \\
 Glorotnormal &               860 &        0.185110 & $<0.001$ \\
     Identity &               360 &        0.186394 & $<0.001$ \\
     Henormal &               N/A &        N/A & $<0.001$ \\
    Heuniform &               N/A &        N/A & $<0.001$ \\
   Orthogonal &               785 &        0.185592  & $<0.001$\\
       Random &               N/A &        N/A & $<0.001$ \\
\end{tabular}
\caption{Convergence metrics reported as averages across 10 runs, with convergence criteria of $\epsilon = 0.005$, and $\alpha = 500$, with N/A recorded when they were not met throughout the training. The p-values reported correspond to the results of single tailed t-test on the loss at the final epoch of the training across the ten runs. Straddled Matrix reached a lower converged loss than all other initialisers.}
\label{figure:allInitSwarm}
\end{subfigure}
\caption{Performance of various initialisers on the Swarm Behaviour dataset.}
\end{figure}

The Swarm Behaviour dataset consists of 2400 features and thus would require the most compression by the autoencoder when being reduced to a latent space with 33 dimensions. The Straddled Matrix initialiser demonstrated a rapid decrease in loss within the first 50 epochs and went on to reach a lower loss than any other initialiser as shown in Figure  \ref{fig:SwarmAll}, with convergence results presented in Figure \ref{figure:allInitSwarm}. The loss at the end of training was highly significantly better than any other method, and, notably, a low loss was achieved in just 20 epochs, 300 epochs earlier than any other method.

\newpage

\section*{Discussion \& Conclusions}
In this article we have introduced and tested a novel weight initialisation technique for Artificial Neural Networks - the \emph{Straddled Matrix}. Straddled Matrix is a generalization of Identity matrix to non-square matrices. This is particularly useful for autoencoders, which is the architecture used in this article. 

The motivation for Straddled Matrix initialiser is the assumption that most relationships in data as well as the transformation performed by this initialiser are linear, therefore minimising reconstruction loss should be an generally smoother process leading to lower error, and faster convergence.
Identity Matrix initialisation does have these same properties for certain architectures, but comes with areas of zeroes whenever the weight matrix is not square.

To verify these hypotheses we have trained a simple fixed-architecture autoencoder on three datasets, ten times each, using seven state-of-the-art initialisation techniques as well as the Straddled Matrix. In all cases except two (out of 21 comparisons) the Straddled Matrix achieved highly significantly better (p-value of $<0.001$) and faster convergence than all of the other initialisers. The two cases where convergence was not significantly better (p-value$>0.001$) were the comparison between the Straddled Matrix and Identity, and Straddled Matrix and Random initialisations on the synthetic dataset we have devised. Yet, even in these two cases Straddled Matrix initialisation converged faster and reached a lower loss  than all other methods (see Figure~\ref{fig:allInitSynthetic}). These results therefore constitute a novel improvement to training autoencoders, saving precious computation time as well as providing more accurate reconstruction results.

We see a large number of opportunities for further investigation regarding the Straddled Matrix initialisation of weights in Artificial Neural Networks.

\begin{enumerate}
    \item Introduce stochasticity to facilitate escapes from local minima by randomisation where the ones are located in the matrix.
    \item Scale the ones in the matrix to achieve even activations across hidden units - a refinement that was shown to provide improvements to other initialisation techniques.
    \item Try different network architectures e.g. feed-forward, convolutional networks and others.
    \item Analyse linear separability of latent space using a support vector machine (SVM). While the results clearly favour Straddled Matrix, they do not directly support our hypothesis. Demonstration of better linear separability of latent space would provide more support for our hypotheses.

\end{enumerate}

\subparagraph{Acknowledgements:} Special thank you belongs to Nikola Loncar and Michał Gnacik for reviewing the code accompanying this article.

\subparagraph{Data \& code availability:} The code for the Straddled Matrix and the experiments to reproduce the results presented here are publicly available on GitHub \url{hhttps://github.com/artefactory-uk/autoencoder-paper}.

\newpage
\bibliography{references}

\begin{thebibliography}{10}

\bibitem{narkhede2022review}
Meenal~V Narkhede, Prashant~P Bartakke, and Mukul~S Sutaone.
\newblock A review on weight initialization strategies for neural networks.
\newblock {\em Artificial intelligence review}, 55(1):291--322, 2022.

\bibitem{Hinton}
G.~E. Hinton and R.~R. Salakhutdinov.
\newblock Reducing the dimensionality of data with neural networks.
\newblock {\em Science}, 313(5786):504--507, 2006.

\bibitem{glorot}
Xavier Glorot and Yoshua Bengio.
\newblock Understanding the difficulty of training deep feedforward neural
  networks.
\newblock In Yee~Whye Teh and Mike Titterington, editors, {\em Proceedings of
  the Thirteenth International Conference on Artificial Intelligence and
  Statistics}, volume~9 of {\em Proceedings of Machine Learning Research},
  pages 249--256, Chia Laguna Resort, Sardinia, Italy, 2010. PMLR.

\bibitem{He}
Kaiming He, Xiangyu Zhang, Shaoqing Ren, and Jian Sun.
\newblock Delving deep into rectifiers: Surpassing human-level performance on
  imagenet classification.
\newblock {\em CoRR}, abs/1502.01852, 2015.

\bibitem{fournier2019empirical}
Quentin Fournier and Daniel Aloise.
\newblock Empirical comparison between autoencoders and traditional
  dimensionality reduction methods.
\newblock In {\em 2019 IEEE Second International Conference on Artificial
  Intelligence and Knowledge Engineering (AIKE)}, pages 211--214. IEEE, 2019.

\bibitem{chandola2009anomaly}
Varun Chandola, Arindam Banerjee, and Vipin Kumar.
\newblock Anomaly detection: A survey.
\newblock {\em ACM computing surveys (CSUR)}, 41(3):1--58, 2009.

\bibitem{an2015variational}
Jinwon An and Sungzoon Cho.
\newblock Variational autoencoder based anomaly detection using reconstruction
  probability.
\newblock {\em Special Lecture on IE}, 2(1):1--18, 2015.

\bibitem{bank2020autoencoders}
Dor Bank, Noam Koenigstein, and Raja Giryes.
\newblock Autoencoders.
\newblock {\em arXiv preprint arXiv:2003.05991}, 2020.

\bibitem{cevora2020fair}
George Cevora.
\newblock Fair adversarial networks.
\newblock {\em arXiv preprint arXiv:2002.12144}, 2020.

\bibitem{kerasAEBlog}
Francois Chollet.
\newblock {Building Autoencoders in Keras}, 2016.

\bibitem{swarmBehaviour}
Shadi Abpeikar and Kathryn Kasmarik.
\newblock Motion behaviour recognition dataset collected from human perception
  of collective motion behaviour.
\newblock {\em Data in Brief}, 47:108976, 2023.

\end{thebibliography}
\bibliographystyle{unsrt}

\end{document}